\documentclass[letterpaper, 10 pt, conference]{ieeeconf}  % Comment this line out if you need a4paper

\IEEEoverridecommandlockouts                              % This command is only needed if
\usepackage{graphics} % for pdf, bitmapped graphics files
\usepackage{epsfig} % for postscript graphics files
\usepackage{mathptmx} % assumes new font selection scheme installed
\usepackage{times} % assumes new font selection scheme installed
\usepackage{amsmath} % assumes amsmath package installed
\usepackage{amssymb}  % assumes amsmath package installed
\usepackage{caption}
\usepackage{subcaption}
\usepackage{graphicx}
\usepackage{amsmath}
\usepackage{amssymb}
\usepackage{booktabs}
\usepackage[pagebackref,breaklinks,colorlinks]{hyperref}
\usepackage{algorithm}
\usepackage{algpseudocode}
\usepackage{amsmath}
\usepackage[capitalize]{cleveref}

\usepackage{xspace}
\newcommand{\etc}{etc.}
\newcommand{\ie}{{\emph{i.e.}},\xspace}
\newcommand{\eg}{{\emph{e.g.}},\xspace}
\newcommand{\etal}{{\emph{et al.}}}

\title{\LARGE \bf
GE-Grasp: Efficient Target-Oriented Grasping in Dense Clutter
}

\author{Zhan Liu, Ziwei Wang, Sichao Huang, Jie Zhou, and Jiwen Lu %<-this % stops a space
\thanks{The authors are with the Beijing National Research Center for Information Science and Technology (BNRist), and the Department of Automation, Tsinghua University, Beijing, 100084, China. (Jiwen Lu is the corresponding author of this paper, {\tt\small {Email: lujiwen@tsinghua.edu.cn.}})
}
}

\begin{document}

\maketitle
\thispagestyle{empty}
\pagestyle{empty}

%%%%%%%%%%%%%%%%%%%%%%%%%%%%%%%%%%%%%%%%%%%%%%%%%%%%%%%%%%%%%%%%%%%%%%%%%%%%%%%%
\begin{abstract}

Grasping in dense clutter is a fundamental skill for autonomous robots. However, the crowdedness and occlusions in the cluttered scenario cause significant difficulties to generate valid grasp poses without collisions, which results in low efficiency and high failure rates. To address these, we present a generic framework called GE-Grasp for robotic motion planning in dense clutter, where we leverage diverse action primitives for occluded object removal and present the generator-evaluator architecture to avoid spatial collisions. Therefore, our GE-Grasp is capable of grasping objects in dense clutter efficiently with promising success rates. Specifically, we define three action primitives: target-oriented grasping for target capturing, pushing, and nontarget-oriented grasping to reduce the crowdedness and occlusions. The generators effectively provide various action candidates referring to the spatial information. Meanwhile, the evaluators assess the selected action primitive candidates, where the optimal action is implemented by the robot. Extensive experiments in simulated and real-world environments show that our approach outperforms the state-of-the-art methods of grasping in clutter with respect to motion efficiency and success rates. Moreover, we achieve comparable performance in the real world as that in the simulation environment, which indicates the strong generalization ability of our GE-Grasp. Supplementary material is available at: https://github.com/CaptainWuDaoKou/GE-Grasp.

\end{abstract}

%%%%%%%%%%%%%%%%%%%%%%%%%%%%%%%%%%%%%%%%%%%%%%%%%%%%%%%%%%%%%%%%%%%%%%%%%%%%%%%%
\section{INTRODUCTION}

Grasping in unstructured environments is a fundamental skill for general purpose robots with numerous applications in manufacturing, logistics, food production, \etc \cite{charalambides2017rapid, 2019Robust}. Although robotic grasping for singulated objects has been widely studied and shown impressive progress \cite{mousavian20196, lenz2015deep}, it is not a common scenario for the target to be completely isolated in realistic applications. Due to the significantly increased uncertainty, effectively grasping objects in clutter is highly desirable while still remaining challenging in robotics.
%  due to the significantly increased grasp uncertainty

\begin{figure}
      \centering
        \vspace{0.2cm}
        \begin{subfigure}{0.92\linewidth}
        \includegraphics[width=1.0\linewidth]{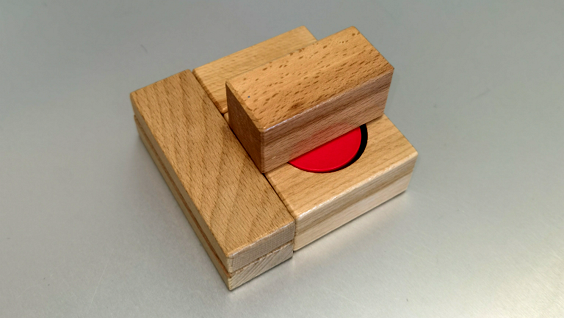}
        \caption{Problem configuration.}
        \label{fig:config}
      \end{subfigure}
    %   \hfill
      \begin{subfigure}{0.94\linewidth}
        \hspace{-0.2cm}
        \includegraphics[width=1.0\linewidth]{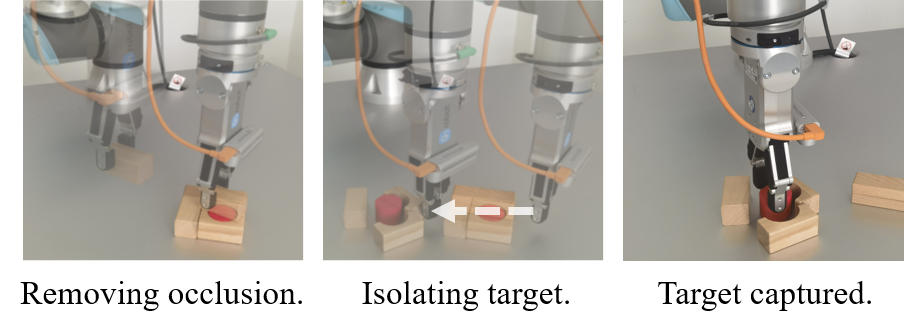}
        \vspace{-0.3cm}
        \caption{Our approach.}
        \label{fig:approach}
      \end{subfigure}
    %   \hfill
        \caption{\textbf{Problem configuration. }The target is a red cylinder, which is occluded by a block on the top and closely surrounded by other blocks. The problem is solved via a sequence of manipulations by our GE-Grasp: 1) a nontarget-oriented grasping to remove the block occluding the target; 2) a pushing breaking the clutter to free the target from the clutter; 3) a target-oriented grasping to pick up the target.}
      \label{fig:config_appr}
      \vspace{-0.6cm}
    \end{figure}

To address this issue, several methods have been proposed for grasping in dense clutter. Adithyavairavan \etal \cite{murali20206} and  Agboh \etal \cite{agboh2018real} attempt to plan collision-free grasps to pick up the target directly, while Zhang \etal \cite{zhang2019multi} propose to grasp the objects in the clutter one by one in a planned order.
% Researchers like Kiatos \etal \cite{kiatos2019robust} and Dogar \etal \cite{2012A} introduce non-prehensile manipulations and intend to achieve the goal via singulation.
Danielczuk \etal \cite{2019Mechanical, 2020X} further investigate approaches for searching and isolating the target object with non-prehensile manipulations like pushing. Recently, the impressive work of Zeng \etal \cite{zeng2018learning} taking advantage of the synergy between pushing and grasping demonstrates that flexibly switching between action primitives with different functions could be a promising solution. Although significant improvement has been made in this area, target-oriented grasping in dense clutter (see \cref{fig:config}) still faces the following challenges. First, the severe occlusion among objects makes it difficult to understand the clutter scenario for valid grasp pose generation. Second, the crowdedness in dense clutter prevents planning the collision-free grasps due to the lack of space.

In this paper, we present the GE-Grasp method to generate robotic manipulations for target grasping in dense clutter. Unlike conventional methods, which make direct regressions on the perception data of a single or multiple objects and output possible grasp poses only, our method designs diverse action primitives flexibly removing occluded objects and present the generator-evaluator architecture for grasp pose generation to efficiently avoid spatial collisions (see \cref{fig:approach}). More specifically, we defined three action primitives including target-oriented grasping for picking up the target directly, pushing to make space for the gripper, and nontarget-oriented grasping to remove the occluded objects around the target. We build the generators to efficiently generate candidates for the operation position of primitive actions by leveraging the spatial correlation test (SCT), which is a series of rules for spatial height detection and searches in the workspace for positions where collision-free force closures could be formed. The evaluators assess a motion candidate by predicting the benefit acquired for target grasping after it is executed in the current scene, where the optimal action is selected by a conditional-greedy policy for robot implementation. Extensive experiments show our GE-Grasp outperforms the state-of-the-art methods of grasping in clutter with respect to motion efficiency and success rates in both simulated and real-world environments. Moreover, our approach achieves comparable performance in the real world as that in the simulated environment, which indicates the strong generalization ability of GE-Grasp.

\section{Related work}
\label{sec:formatting}

\begin{figure*}[t]
  \centering
  \vspace{0.1cm}
   \includegraphics[width=1\linewidth]{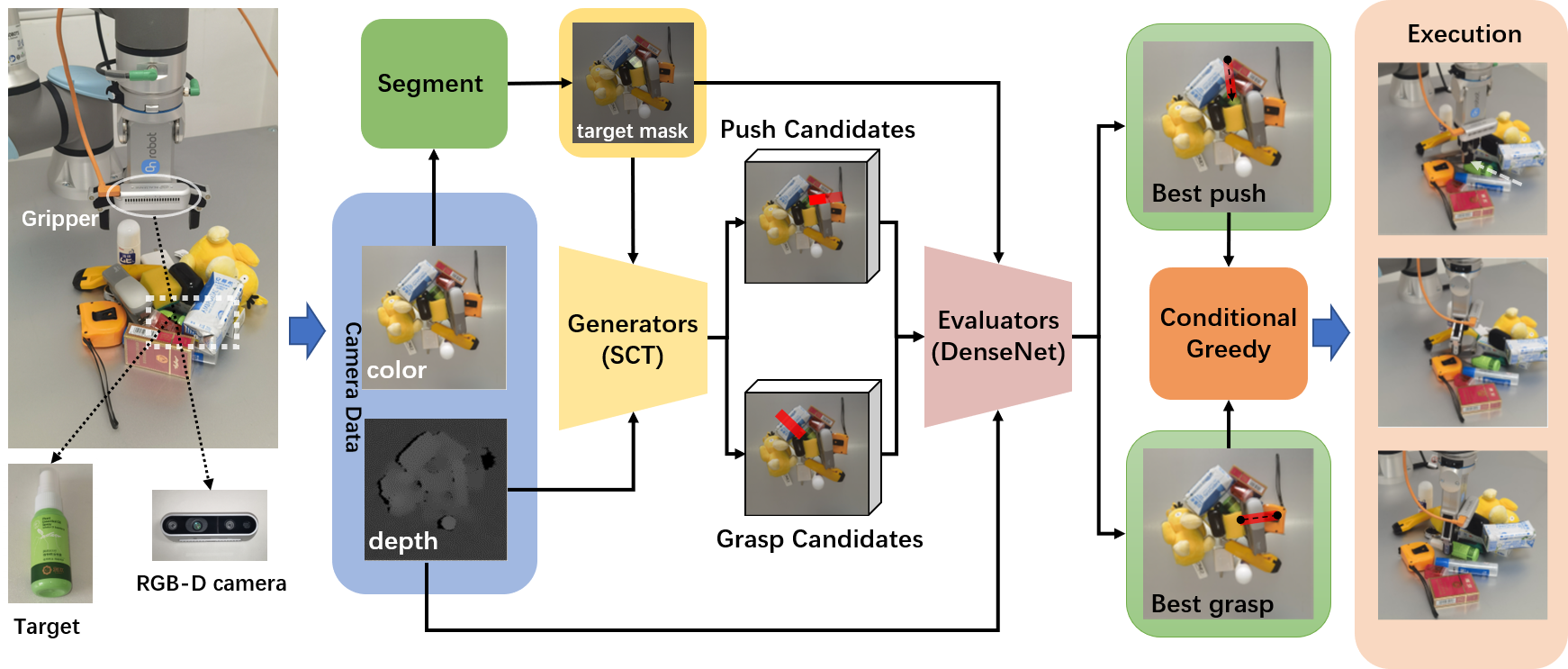}
   \caption{\textbf{Overview.} Our robot manipulates on a tabletop observed by an RGB-D camera from a fixed position and angle. The segmentation module takes in the RGB image and outputs a mask indicating the target object. Multiple motion candidates including pushing, nontarget-oriented grasping and target-oriented grasping are provided by the generators based on the depth heightmap, which are fed into the evaluators together with the target mask and depth heightmap for quality assessment. The optimal action for execution is chosen from the best push and best grasp through a conditional greedy policy.}
   \label{fig:overview}
   \vspace{-0.3cm}
\end{figure*}

\textbf{Grasping: }Classical solutions for robotic grasping aim to find stable force closures by explicitly modeling the physical dynamics of both the objects and the gripper \cite{2012An,7139625}. However, these approaches heavily rely on strong assumptions that usually do not hold in practice and the prior knowledge of objects that is hard to acquire in real-world settings. In recent years, considerable advancements have been witnessed in data-driven methods due to the application of deep learning techniques for robotic vision \cite{lenz2015deep,pinto2016supersizing}, which enable robots to learn successful grasps with enhanced capabilities compared to hand-crafted methods. Since the real-world scenarios usually contain multiple objects with complex interactions, the data-driven methods that focus on dealing with isolated objects \cite{lu2020planning, mousavian20196} acquire significantly decreased performance in practice. To address this, some researchers decompose the problem of grasping in dense clutter into sequentially capturing isolated objects. Zhang \etal \cite{zhang2019multi} propose the grasp-only strategy that consists of simultaneous robotic grasping detection and manipulation relationship reasoning, where the target is finally captured by removing covered objects in the learned order. However, the grasping-only strategy fails to handle adversarial cases appearing in practice with high frequencies, where valid grasps may not exist due to the object occlusion and crowdedness.

\textbf{Pushing: }The study of pushing can be traced back to the early days of robotic manipulation for tasks like driving an object to a specific pose and position \cite{hermans2012guided, lynch1996stable}. A large amount of these methods for robotic pushing are model-based and require prior knowledge of physical properties (\eg shapes, weights, friction, \etc) \cite{mason1986mechanics, 2012Interactive}. Recent works have explored end-to-end learning approaches to map visual observations into pushing planning and achieved promising results \cite{zhou2016convex, clavera2017policy, bauza2017probabilistic}. Eitel \etal \cite{eitel2020learning} select favorable push actions in clutter to separate the target from unknown objects, and Zhou \etal \cite{zhou2016convex} aim to learn the force-motion dynamics with a polynomial model. However, these methods mainly focus on planning stable pushing policies without collaborating with other actions and limit the practicality of robotic manipulation that requires complicated interactions.

\textbf{Pushing with grasping: }Effective non-prehensile manipulations like pushing can singulate the target object by breaking the structure of the clutter, where enough space is prepared for the gripper. Ignasi \etal \cite{clavera2017policy} explore the model-free planning that drives the target object to the position suitable for pre-designed grasping algorithms. To jointly select the optimal grasping poses and pushing positions, Zeng \etal \cite{zeng2018learning} present a simultaneous learning method for complementary pushing and grasping policies from scratch through self-supervised trials and errors. Yang \etal \cite{yang2020deep} further introduce a segmentation module to flexibly denote the target, where the decision of pushing and grasping is determined by predicted values based on domain knowledge. However, these methods still face \ the challenges of occlusion and crowdedness due to the limited action primitives and lack of collision-free planning. We follow the task settings illustrated in \cite{yang2020deep} but propose a completely different architecture. Our approach applies collaborative pushing and grasping actions in a data-driven manner without requirements of prior knowledge or precise physical assumptions.

\section{Approach}

We first introduce the overall pipeline of our GE-Grasp. Then we detail the diverse action primitives and the generator-evaluator architecture. Finally, the procedures of training and testing including dataset collection, learning objective and beneficial techniques are demonstrated.

%-------------------------------------------------------------------------
\subsection{Overall Pipeline}

Grasping objects in dense clutter is highly desirable in realistic applications, while still remaining challenging due to the uncertainty in the complex environment. Given the clutter composed of multiple objects with the target inside, a robot aims to pick up the target object which is placed in an arbitrary pose with at least part of it visible. The obstacle objects can be either graspable or ungraspable, and the number of manipulations is also expected to be minimized with high efficiency. However, significant crowdedness and occlusion in the cluttered scenario cause difficulties to generate valid grasp poses without collisions.

To address this issue, we present the GE-Grasp method that generates collision-free grasp poses via sequential pushing and grasping in a collaborative way, where \cref{fig:overview} illustrates the pipeline of the presented GE-Grasp. First, RGB-D images in the top-down view are collected to observe the workspace to perceive color and depth information. The RGB images are fed forward to the pre-trained semantic segmentation module to identify the target location via the predicted masks, and the depth heightmaps are leveraged to efficiently produce candidates of pushing and grasping by the generators with SCT. Finally, the evaluators assess all candidates by predicting their benefits on target grasping in the current scene, where we utilize a conditional greedy policy to select the optimal action candidate for robot implementation. As a result, our GE-Grasp reduces the crowdedness and occlusion in clutter by pushing and removing occluded objects for ungraspable targets, which generates valid grasp poses in complex cases for picking up the target objects.

\subsection{Primitive Actions}

Conventional methods utilize five-dimensional bounding box \cite{2013Deep, levine2016end, eitel2015multimodal} to represent robotic grasps. This approach describes a grasp by a rectangle with a specified position, size, and orientation, which is implemented by a parallel plate gripper. However, the conventional representations fail to acquire promising performance due to the neglect of rich visual information. Inspired by \cite{mousavian20196}, we represent the robot actions by masks with the guidance of visual cues. To generate valid grasp poses in adversarial cases, we define diverse action primitives including pushing, nontarget-oriented grasping, and target-oriented grasping.

\textbf{Pushing: }Effective pushing contributes to breaking structured clutter and separating the target object from others. The primitive action of pushing is that the closed gripper moves along a straight line parallel to the tabletop in order to push objects with fingertips. Pushing can be represented by a rectangular mask revealing the starting position and the orientation. As shown in \cref{fig:push-2}, in the rectangular mask, pixels in the first and the second halves of the pushing route are assigned with 0.5 and 1.0 respectively. Therefore, the mask representing pushing indicates the area swept by the gripper in the workspace with the starting point and orientation.

\textbf{Nontarget-oriented grasping:} Although pushing breaks the clutter structure to make space for the gripper, accurately predicting the consequent changes in dense clutter is difficult. To eliminate the occlusion in a more fine-grained way, we present nontarget-oriented grasping that deterministically removes the obstacles covering or surrounding the target. As shown in \cref{fig:grasp-2},  nontarget-oriented grasping is represented as a rectangular all-ones mask, which demonstrates the position and orientation of the fully opened gripper.

\textbf{Target-oriented grasping: }Target-oriented grasping is described in the same way as nontarget-oriented grasping but aims to pick up the target object directly. Currently, target-oriented grasping is executed as the last step after the obstacles are cleared by the other two action primitives.

%-------------------------------------------------------------------------

\subsection{Generators}

The generators are designed to produce valid motion candidates via heuristic methods. Since we assume that actions close to the objects exert great influence, the actions are generated within the region of interest, which is a small square area centered on the target. We regard every pixel in the region of interest as a source pixel from which action with random orientation is generated. Although the number of the generated actions is large, the push generator and the grasp generator are designed to efficiently select collision-free motion candidates for the evaluators by SCT.

\begin{figure}
  \centering
  \vspace{0.2cm}
  \begin{subfigure}{0.32\linewidth}
    \includegraphics[width=1.0\linewidth]{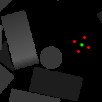}
    \caption{Height detection.}
    \label{fig:push-1}
  \end{subfigure}
%   \hfill
  \begin{subfigure}{0.32\linewidth}
    \includegraphics[width=1.0\linewidth]{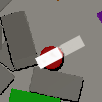}
    \caption{Push mask.}
    \label{fig:push-2}
  \end{subfigure}
%   \hfill
  \begin{subfigure}{0.32\linewidth}
    \includegraphics[width=1.0\linewidth]{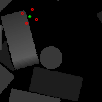}
    \caption{Invalid push.}
    \label{fig:push-3}
  \end{subfigure}
    \caption{\textbf{Generating push candidates}. (a) demonstrates the SCT for push candidates, where height of the four red points should be lower than the target. (b) illustrates the candidate represented by a push mask. (c) shows an example of the invalid push that leads to unwanted collisions.}
  \label{fig:push_gen}
  \vspace{-0.3cm}
\end{figure}

\textbf{Push generator: }The source pixel denotes the starting position of pushing in one of three available directions including facing the target and deflecting 22.5° to the left and right, which is performed with the fingertips of the closed gripper. Enough space should be prepared at the starting position for the closed gripper in order to avoid unwanted collisions. Therefore, the SCT for the action primitive of pushing is that the value of the source pixel in the heightmap should be smaller than the pixel covered by the target (see \cref{fig:push-1}), which indicates that the gripper can be placed vertically to the starting position without any collisions.

However, since we traverse the candidates formed from every pixel in the region of interest, actions generated from adjacent pixels may fall into similar behaviors and lead to the same result. The massive redundancy in the candidates reminds us that the number of candidates can be reduced by sampling, so as to reduce computational costs. Considering that the generation of motion candidates is highly related to the altitudes in the scenario, the imbalanced distribution of objects in the scenario may result in an imbalanced distribution of pushing candidates, which extremely limits the expressiveness of pushing. We divide the workspace into four quadrants with the target center as the origin and randomly sample a maximum of 25 candidates in each quadrant, a maximum of 100 candidates in total, which achieves a good balance between including the optimal action and reducing the amount of computation. Through this method, candidates in any direction of the target can be obtained.

\begin{figure}
  \centering
  \vspace{0.2cm}
  \begin{subfigure}{0.32\linewidth}
    \includegraphics[width=1.0\linewidth]{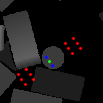}
    \caption{Height detection.}
    \label{fig:grasp-1}
  \end{subfigure}
%   \hfill
  \begin{subfigure}{0.32\linewidth}
    \includegraphics[width=1.0\linewidth]{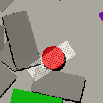}
    \caption{Grasp mask.}
    \label{fig:grasp-2}
  \end{subfigure}
%   \hfill
  \begin{subfigure}{0.32\linewidth}
    \includegraphics[width=1.0\linewidth]{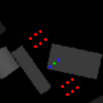}
    \caption{Empty grasp.}
    \label{fig:grasp-3}
  \end{subfigure}
    \caption{\textbf{Generating grasp candidates}. (a) depicts the SCT for grasp candidates, where the two blue points should be higher than the red points. (b) shows the candidate represented by a grasp mask. (c) shows an example of the empty grasp that passes the SCT but leads to grasping failures.}
  \label{fig:grasp_gen}
  \vspace{-0.3cm}
\end{figure}

\textbf{Grasp generator: }In the grasp generator, the source pixel stands for the middle position of the parallel jaw grasp in the top-down view. Moreover, $16$ types of orientations can be selected for grasp candidates generated at source pixels, where the difference between adjacent orientations is $22.5^{\circ}$. The grasp generator is designed to find potential collision-free force closure solutions in the workspace without analytical modeling of the objects and the robot. Since we employ the parallel jaw as the gripper, a top-down grasp can be completed when a sizable height difference between the finger and the gripper center exists. Therefore, the SCT for action primitives of nontarget-oriented grasping and target-oriented grasping is that the value of the source pixel in the heightmap is required to be larger than the pixel covered by fingers (see \cref{fig:grasp-1}). The accurate height difference required is calculated with reference to the trajectory of the fingertips when the gripper is closing. The SCT enables a potential force closure while avoiding unwanted collisions during gripper closing. In order to enhance the efficiency of grasp generation and the expressiveness of action candidates, the same sampling method is applied as in the push generator.

%-------------------------------------------------------------------------
\subsection{Evaluators}

The candidates include most effective action primitives that benefit capturing the target, but there are still a few empty grasps with failure trails, as shown in \cref{fig:grasp-3}. Hence, an evaluator is expected to distinguish the successful and failed candidates. The evaluators assess the motion candidates by predicting the benefits of the actions for target grasping and choose the optimal one. We build a push evaluator and a grasp evaluator respectively, which are both modeled by DenseNet-121 \cite{huang2017densely}. The input for each evaluator consists of the current heightmap, target mask, and a push/grasp mask that represents the motion candidate, and the evaluators output the predicted value of the input action candidates.

The action candidates with high scores are regarded to be beneficial, and only actions whose score is higher than the pre-defined threshold are considered to be effective. Grasping is preferred to pushing in our GE-Grasp since grasping brings more deterministic consequences other than uncertainty. Therefore, we propose a conditional greedy policy for implemented action selection. The best grasp candidate is selected if the score surpasses the threshold, indicating the effectiveness of the preferred grasp actions. Otherwise, the candidate with the highest score is chosen to be executed.

% \begin{equation}
%     a=\pi_c(p, g) = \left \{
%         \begin{array}{rcl}
%         g& {v_g \geq thr& or&v_g \geq v_p}\\
%         p& {else}\\
%         \end{array} \right.
%         \label{eq:important}
% \end{equation}

%-------------------------------------------------------------------------
\subsection{Training and Testing}

In this section, we introduce the details of dataset collection, the learning objective for model training, and the beneficial techniques during testing. Training data for the evaluators are collected in the simulation environment of V-REP (a popular robot simulation platform), where the robot randomly performs actions provided by the generators.

\begin{figure}[t]
  \centering
  \vspace{0.2cm}
  \begin{subfigure}{0.32\linewidth}
    \includegraphics[width=1.0\linewidth]{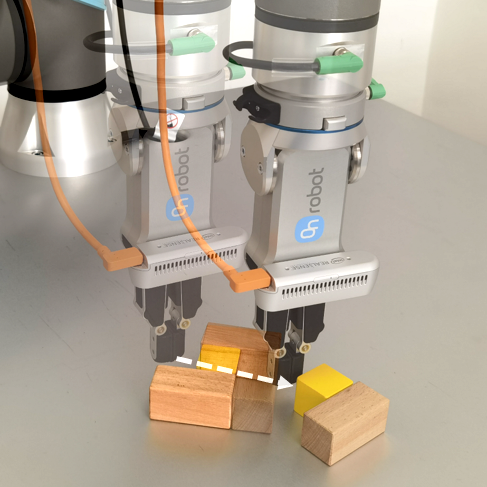}
    % \caption*{\footnotesize Crowdedness reduced, value = 1}
    \caption*{\footnotesize value = 1}
    % \caption{\footnotesize }
    \label{fig:label-1}
  \end{subfigure}
%   \hfill
  \begin{subfigure}{0.32\linewidth}
    \includegraphics[width=1.0\linewidth]{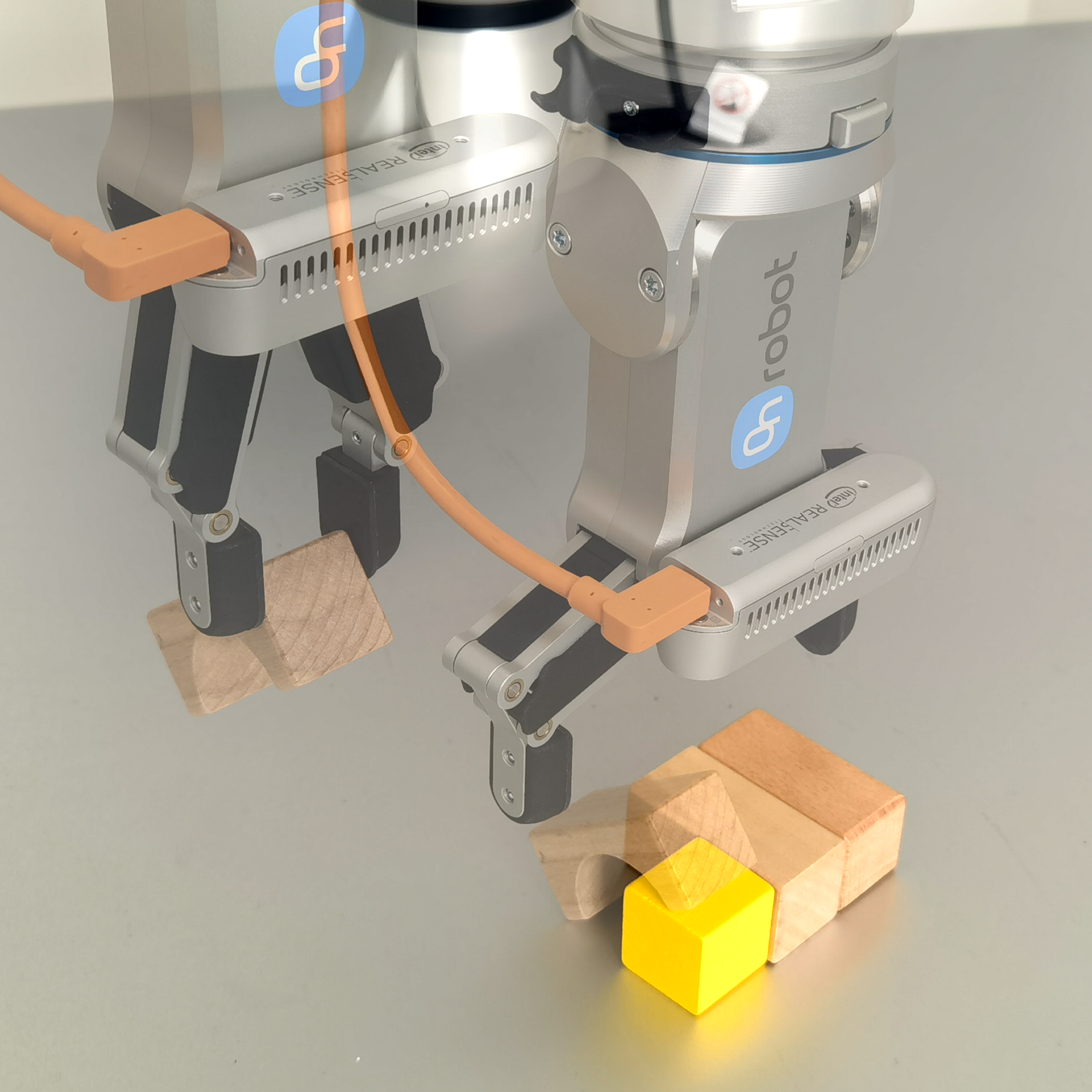}
    % \includegraphics[width=1.0\linewidth]{05-2.png}
    % \caption*{\footnotesize Occlusions reduced.}
    \caption*{\footnotesize value = 1}
    \label{fig:label-2}
  \end{subfigure}
%   \hfill
  \begin{subfigure}{0.32\linewidth}
    \includegraphics[width=1.0\linewidth]{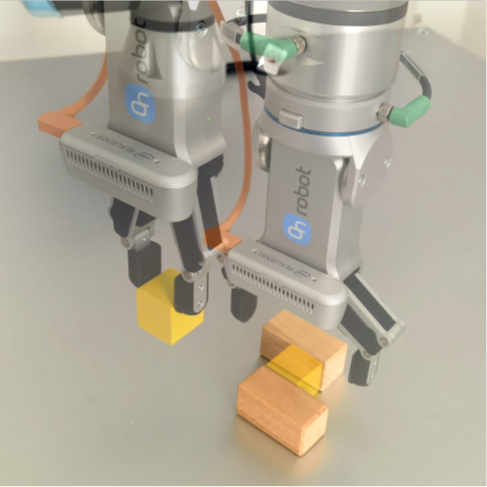}
    % \caption*{\footnotesize Target grasped.}
    \caption*{\footnotesize value = 2}
    \label{fig:label-3}
    \vspace{-0.2cm}
  \end{subfigure}
  \caption{\textbf{Ground-truth value assigning. }Ground-truth values of actions are assigned differently according to the consequences after execution. Actions that reduce crowdedness and occlusions in the clutter are regarded as effective and assigned with 1, while target-oriented grasping that successfully picks up the target is assigned with 2.}
  \label{fig:label}
  \vspace{-0.5cm}
\end{figure}

An action can be considered to have reached the ultimate goal, effective or meaningless, based on its contribution to capturing the target object, and is therefore assigned a value of 2, 1, or 0. Action with a higher value is always preferred to be performed. We regard the action of pushing as effective if the occlusion over the target decreases or the space around the target increases by a certain threshold. To quantitatively represent the occlusion and crowdedness, domain knowledge of the workspace is introduced \cite{yang2020deep}. We compare the target masks predicted by the segmentation module before and after the action, which are denoted as $m_t$ and $m_{t’}$ respectively. If the number of pixels covered by the target in $m_{t’}$ is 20\% more than that in $m_t$, we regard the occlusion to decrease significantly. For crowdedness, we construct the mask of target border $m_b$ by expanding the target mask, and the target border occupancy value $o_b$ is defined as the number of pixels with height above the ground, which indicates the amount of space around the target object occupied by obstacles. The decrease in $o_{b}$ after performing an action shows an increase in free space around the target, \ie a decrease in crowdedness. Finally, as shown in \cref{fig:label}, samples with effective pushes are assigned with 1 and otherwise 0.

Effective grasps can either capture the target directly or remove objects occluding the target. We assign the value as 2 for the grasp that successfully picked up the target, and the value is assigned with 1 for that attempting towards nontarget objects and proven to be effective (see \cref{fig:label}). Samples that result in grasp failures or provide no benefits for target grasping in the current scenario are assigned by 0. We use the same evaluation criteria as that in the value assignment of the push to determine whether a grasp towards nontarget objects is effective or not.

% We collected a dataset for pushing and grasping respectively by randomly executing actions provided by the generators in V-REP, and each dataset contains 4400 samples.

We collected a dataset for pushing and a dataset for grasping separately by randomly executing actions provided by the two generators in V-REP, and the two datasets are of the same size with 4400 samples. Each sample consists of a heightmap of the current scene, a target mask provided by the segmentation module, and a mask of the action primitive being performed. The samples are labeled by evaluating the benefits to the goal of target grasping in dense clutter. The evaluators are optimized offline with batch mode training utilizing the collected dataset by minimizing the error between predicted score and ground-truth values via L1 loss.

As no physical boundaries encompass the workspace, objects may lie on the edge with incomplete perceptions. To enable the target on edge to be graspable, a deterministic pushing will be performed, where the target is pushed back to the workspace for complete information acquisition. Meanwhile, we force the robot to execute pushing if it fails to grasp twice consecutively to avoid trivial solutions.

%-------------------------------------------------------------------------
\section{Experiments}

We conduct extensive experiments in both simulated and real-world environments to evaluate our GE-Grasp. The goal of the experiments is to verify that 1) the generators can provide multiple valid collision-free primitive action candidates, 2) the evaluators are capable of choosing the optimal action for target grasping, 3) motion efficiency of target-oriented grasping in dense clutter is improved sizably and 4) our GE-Grasp acquires strong generalization ability to novel objects in the real world. In simulation experiments, we followed the same settings as in \cite{yang2020deep} for a fair comparison. In real-world experiments, we modify our model to enable the robot to explore the target in dense clutter, where a deterministic pushing is added before at the beginning of the action sequences. Furthermore, we also evaluate our method with the same task in completely novel objects.

%-------------------------------------------------------------------------
\subsection{Implementation Details}

The workspace is a $0.448^2 m^2$ area on the tabletop, and the visual observations are converted into 224$\times$224 pixel resolution RGB and depth images. In the visual input, each pixel represents a 2$\times$2 $mm^2$ vertical column in the 3D space. The region of interest for candidate generation is a $0.2^2 m^2$ square centered on the target, which contains an area of 100$\times$100 pixels projected to the heightmap. Pushing is represented with a mask of 62$\times$12 pixels according to the gripper size and the motion length, and a grasp mask is set to 60$\times$12 pixels considering the size of the opened gripper. The threshold in our conditional greedy policy is set to 1.0 and the height difference threshold of the SCT for push and grasp generation is 15 $mm$ and 25 $mm$, respectively.

Our hardware configurations include an Intel i5-8500 CPU with an NVIDIA GeForce GTX 1080Ti GPU for acceleration. The evaluator networks are trained with a fixed learning rate of $10^{-4}$ and weight decay $2^{-5}$ by the SGD optimizer. We utilize the pretrained Light-Weight RefineNet \cite{2018Light} as the segmentation module for target annotation.

\begin{figure}[t]
  \centering
  \vspace{0.2cm}
  \includegraphics[width=1.0\linewidth]{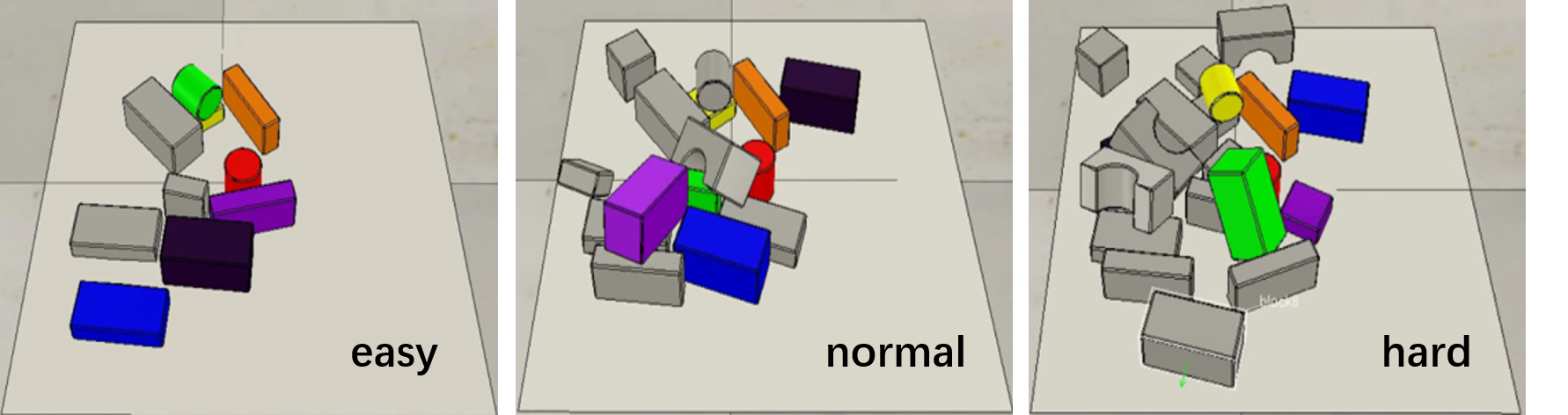}
  \caption{\textbf{Test cases in the random clutter task.} The easy, normal, and hard case contains 10, 15, and 20 randomly generated and placed blocks respectively. We assign one of the colored blocks as the target in each round.}
  \label{fig:rand_clu}
  \vspace{-0.4 cm}
\end{figure}

%-------------------------------------------------------------------------
\subsection{Baseline Methods}
The baseline methods also learn collaborative pushing and grasping to deal with the problem of robotic grasping in dense clutter, which is introduced in detail as follows.

MASK-VPG \cite{zeng2018learning, yang2020deep} is an end-to-end model which takes in visual observations and outputs pixel-wise Q-maps for pushing and grasping. Pushing is executed in a small region around the target while grasping in the area covered by the target mask. The model is trained with reinforcement learning algorithms, and the action with the highest Q-value within the constrained action areas will be executed.

Grasping-Invisible (GI) \cite{yang2020deep} introduces a segmentation module to annotate the target object for optimal action selection of grasping in dense clutter. The critic predicts the Q-value predictions for pushing and grasping, and a classifier-based coordinator incorporates the predicted Q-value with domain knowledge to coordinate pushing and grasping for the detected target. For scenarios with no detected target, a Bayesian-based explorer will search for the target.

%-------------------------------------------------------------------------

\subsection{Evaluation Metrics}
The experiments are executed for 30 and 10 runs in simulation and real-world tests respectively. As our goal is to successfully pick up the target with minimum actions, we evaluate the average performance with motion efficiency and success rate with respect to efficiency and effectiveness:

Motion efficiency (\textbf{ME}) is defined as the number of actions performed before completion divided by the number of target objects, which represents the average number of motions executed per target. Success rate (\textbf{SR}) describes the ratio of successful grasps to the overall trials. A task is successfully completed if the robot captures the target within 5 motions in simulation or 15 motions in the real world.

%-------------------------------------------------------------------------
\subsection{Simulation Experiments}

\begin{figure}[t]
  \centering
   \includegraphics[width=1.0\linewidth]{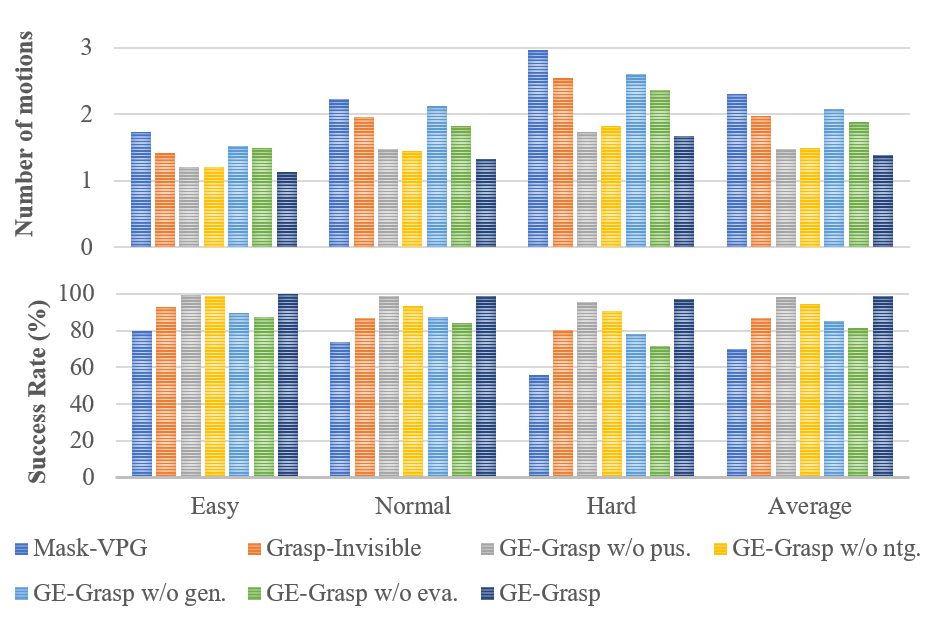}
   \caption{\textbf{Performance of grasping in random clutter with different hardness.} The GE-Grasp w/o pus., ntg., gen., and eva. stands for the variant of GE-Grasp without the pushing primitive, nontarget grasping primitive, generators, and evaluators respectively. Our approach shows a high effectiveness by achieving a task success rate of 98.8\% (bottom) with 1.38 motions in average (top). }
   \label{fig:rand_perf}
  \vspace{-0.5cm}
\end{figure}

The simulation environment is set up in V-REP, where a UR5 robot and an RG2 gripper with Bullet Physics 2.83 are applied for dynamics and V-REP's internal inverse kinematics module is leveraged for robot motion planning. Visual information of the scene is captured from an RGB-D camera statically mounted 0.5 $m$ above the workspace. Despite the baseline methods and GE-Grasp, the no-pushing variant of GE-Grasp which only utilizes the grasping module (\ie GE-Grasp w/o pushing) is also evaluated in simulation.

\begin{table}[t]
  \normalsize
  \centering
  \vspace{0.2cm}
  \begin{tabular}{lccc}
    \toprule
    Method & Success Rate (\%) & Motions\\
    \midrule
    MASK-VPG \cite{yang2020deep, zeng2018learning} & 70.2 & 2.59$\pm$0.47\\
    GI \cite{yang2020deep} & 86.7 & 1.97$\pm$0.32\\
    \hline
    GE-Grasp w/o pus. & 98.1 & 1.47$\pm$0.15\\
    GE-Grasp w/o ntg. & 94.2 & 1.49$\pm$0.18\\
    GE-Grasp w/o gen. & 85.0 & 2.08$\pm$0.23\\
    GE-Grasp w/o eva. & 81.2 & 2.14$\pm$0.26\\
    \hline
    \textbf{GE-Grasp} & \textbf{98.8} & \textbf{1.38$\pm$0.12}\\
    \bottomrule
  \end{tabular}
  \caption{The success rate and the average number of motions for grasping in the random clutter.}
  \label{tab:rand_perf}
  \vspace{-0.2cm}
\end{table}

\begin{figure}[t]
  \centering
  \includegraphics[width=1.0\linewidth]{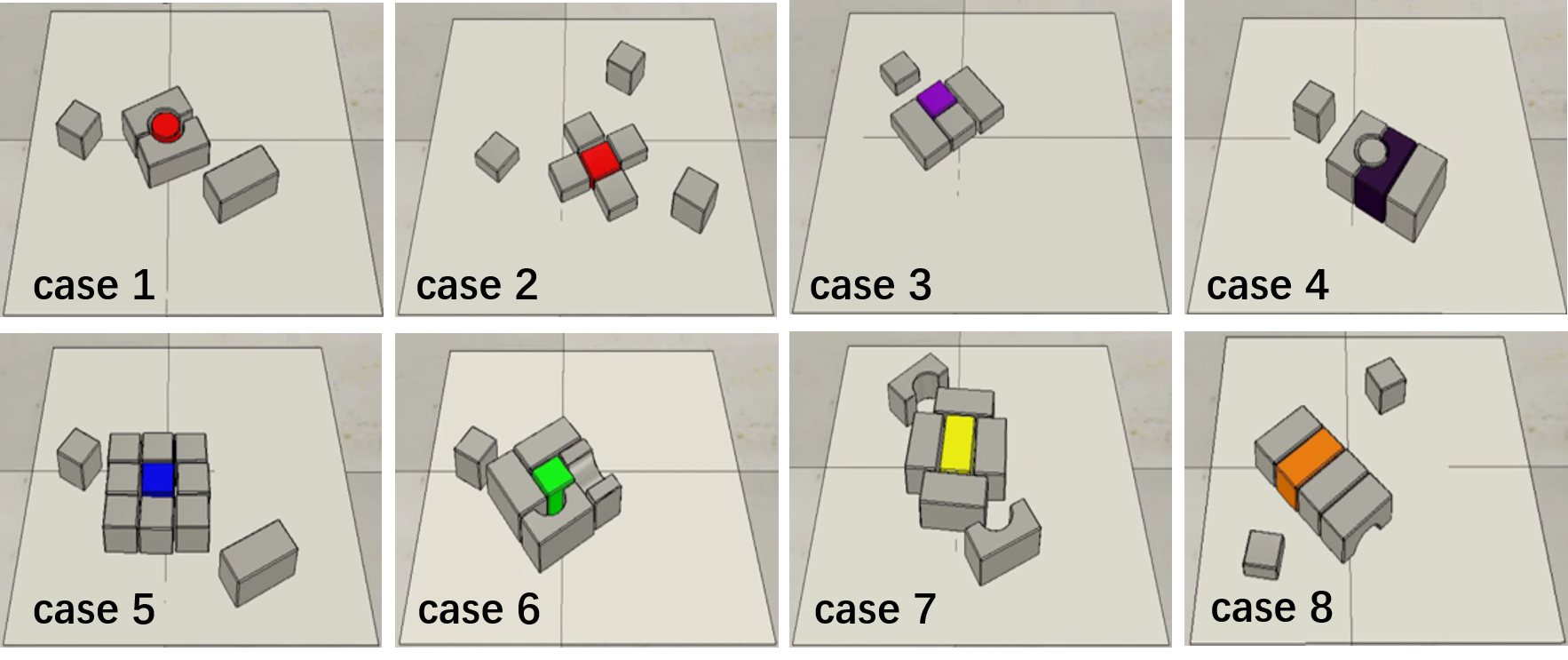}
  \caption{\textbf{Test cases with the challenging clutter.} We set up 8 challenging test cases where objects are adversarially arranged and directly grasping the target is impossible.}
  \label{fig:chal_clu}
  \vspace{-0.5cm}
\end{figure}

\subsubsection{Random clutter}

For random clutter settings, blocks with different colors and shapes are randomly dropped on the workspace. In fact, the target object is difficult to be completely blocked, so it is at least partially visible. In each trial, one of the blocks is assigned as the target object while the others are regarded as obstacles. We set up test cases of random clutter containing 10, 15, and 20 objects respectively, which represent the easy, normal, and hard scenarios (See \cref{fig:rand_clu}). We performed 30 runs on each case, where the results are shown in \cref{fig:rand_perf} and \cref{tab:rand_perf}. Our GE-Grasp outperforms all the other compared strategies remarkably in both success rate and motion efficiency. Overall, our approach achieves a 98.8\% task success rate with 1.38 motions to pick up a target on average. Meanwhile, the no-pushing variant achieves comparable performance which also exceeds the other two baselines. Without the utilization of pushing, the performance of the variant decrease by 0.7\% in task success rate and increase the number of motion by 0.1 in efficiency. Although target information is considered in MASK-VPG and GI, they still face the challenges of occlusion due to the limited action primitives and fail to generate collision-free grasp poses because of the neglect of spatial constraint. Our GE-Grasp generates valid action candidates by designing diverse action primitives and SCT, outperforming the state-of-the-art method GI by 12.1\% (98.8\% vs 86.7\%) in task success rate and 42.8\% (1.38 vs 1.97) in motion efficiency.

To investigate the importance of different components of our framework, we remove nontarget grasping, generators, and evaluators from the original GE-Grasp, respectively, to create a series of variants. The performance of these variants is also tested on the random clutter cases with results shown in \cref{fig:rand_perf} and \cref{tab:rand_perf}. Without nontarget grasping, the \textbf{ME} and \textbf{SR} fell by 7.4\% and 4.6\% respectively, demonstrating the benefit of diversity of action primitives. The \textbf{ME} and \textbf{SR} fell by 50.4\% and 13.8\% with frequent collisions when replacing the generator with a random grasp sampler. The evaluator aims to choose the best action by assessing the benefits of action candidates, and we measure the importance by replacing the evaluators with random sampling. Although there are few collisions, the robot keeps performing actions that are irrelevant to the target capturing, resulting in a significant decrease in \textbf{ME} by 55.7\% and \textbf{SR} by 17.6\%.

\subsubsection{Challenging clutter}
\begin{figure}[t]
  \centering
   \includegraphics[width=1\linewidth]{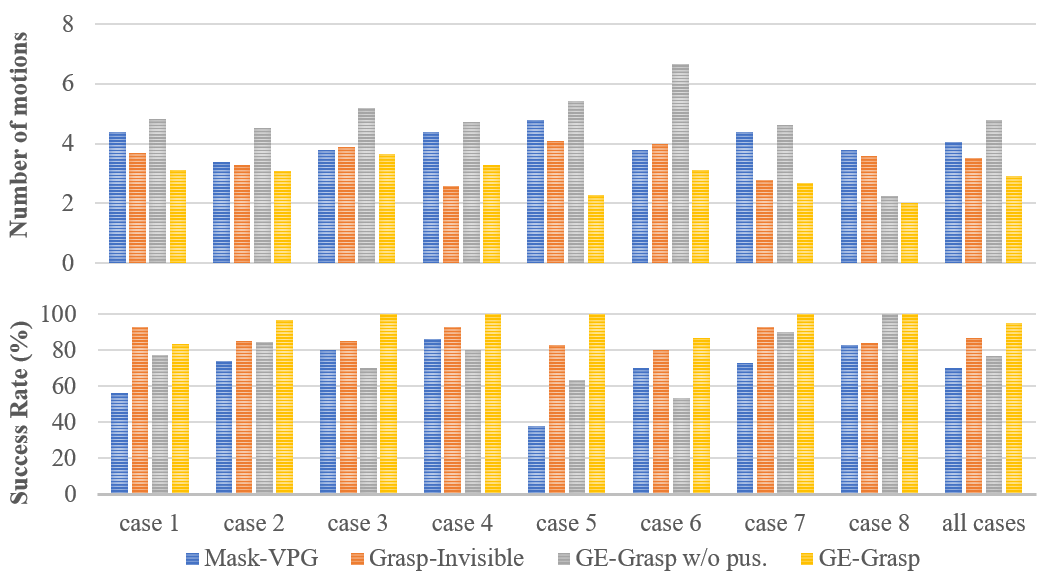}
   \caption{\textbf{Performance of grasping in challenging clutter with different cases.} The plot clearly demonstrates the effectiveness of our approach which achieves a task success rate of 95\% (bottom) with 3.0 motions in average (top).}
   \label{fig:chal_perf}
  \vspace{-0.8cm}
\end{figure}

To further verify the effectiveness and efficiency of our approach on more adversarial cases, we evaluate GE-Grasp on 8 challenging cases provided by \cite{yang2020deep}, where adversarial arrangements are designed to ensure that direct grasping towards the target is infeasible (see \cref{fig:chal_clu}). \cref{fig:chal_perf} and \cref{tab:chal_perf} illustrate the performance of different methods in the challenging clutter test, and the evaluation metric remains the same as that in random clutter. The effectiveness and the efficiency of the no-pushing variant drop significantly in the challenging cases due to the limited action primitives and the lack of the ability to efficiently break structured patterns. On the contrary, the complete GE-Grasp improves the success rate by 7.5\% (95.0\% vs. 87.5\%) and decreases the number of actions by 0.5 (3.0 vs. 3.5) respectively compared to the state-of-art method GI.

MASK-VPG does not behave smart enough to coordinate appropriately between pushing and grasping, and repeatedly executes unnecessary pushing when the target is already graspable. GI alleviates this problem by introducing a coordinator module, however, neither of the two baseline methods can effectively avoid potential collisions, especially when there are multiple objects densely stacked together, which leads to failures of grasping. The reason may be that the pixel-wise mapping network adopted is not sensitive to the details that cause collisions. Our GE-Grasp first provides multiple collision-free action candidates and then selects the optimal one through the evaluators to ensure the success rate of execution while improving task efficiency. By observing the behavior of the robot, we discover that effective non-target grasping considerably reduces the crowdedness and occlusion around the target object so that the robot has a good chance to capture the target in the next step. Pushing may introduce some uncertainties (\eg the sliding or rolling of objects), making it less efficient. However, in some scenarios where grasping cannot be performed directly, the ability of pushing to break the scene structure is indispensable.

\begin{table}[t]
%   \small
  \normalsize
  \centering
  \vspace{0.2cm}
  \begin{tabular}{lccc}
    \toprule
    Method & Success Rate (\%) & Motions\\
    \midrule
    MASK-VPG \cite{yang2020deep, zeng2018learning} & 70.2 & 4.06$\pm$0.83\\
    GI \cite{yang2020deep} & 87.5 & 3.51$\pm$0.90\\
    GE-Grasp w/o pus. & 77.2 & 4.79$\pm$0.97\\
    \textbf{GE-Grasp} & \textbf{95.0} & \textbf{3.02$\pm$0.52}\\
    \bottomrule
  \end{tabular}
  \caption{The success rate and the average number of motions for grasping in the challenging clutter. The no-pushing variant struggles in dealing with adversarial cases where valid grasp poses are hard to acquire.}
  \label{tab:chal_perf}
  \vspace{-0.5cm}
\end{table}

%-------------------------------------------------------------------------
\subsection{Real-world Experiments}

\begin{figure}[t]
  \centering
  \vspace{0.15cm}
  \begin{subfigure}{0.24\linewidth}
    \includegraphics[width=0.9\linewidth]{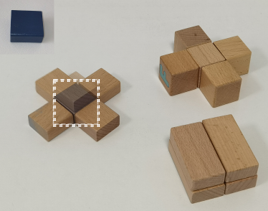}
    \caption{blue cube.}
    \label{fig:short-a}
  \end{subfigure}
  \hfill
  \begin{subfigure}{0.24\linewidth}
    \includegraphics[width=0.9\linewidth]{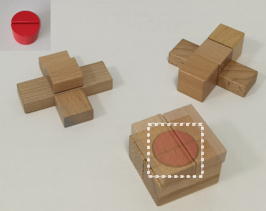}
    \caption{red cylinder.}
    \label{fig:short-b}
  \end{subfigure}
  \hfill
  \begin{subfigure}{0.24\linewidth}
    \includegraphics[width=0.9\linewidth]{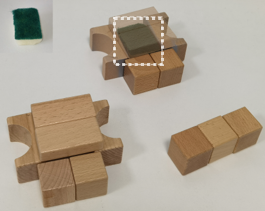}
    \caption{sponge.}
    \label{fig:short-c}
  \end{subfigure}
  \hfill
  \begin{subfigure}{0.24\linewidth}
    \includegraphics[width=0.9\linewidth]{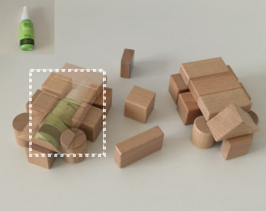}
    \caption{spray bottle.}
    \label{fig:short-d}
  \end{subfigure}

  \caption{\textbf{Test cases of the ``Grasping the Invisible'' task in the real-world environment.} The target is buried in the clutter and invisible initially.}
  \label{fig:invi_set}
  \vspace{-0.4cm}
\end{figure}

In order to evaluate the effectiveness and efficiency of our GE-Grasp, we conducted experiments in the real world where the model is trained with the data collected in the simulation environment. We apply a UR5e robot arm with an RG2 gripper which implements operations on the desktop in front of them. Due to the space limitation, the Realsense D435 camera is mounted on the gripper and captures visual information in a fixed position and orientation. In different experimental settings, the targets are toy blocks or everyday objects with different shapes and colors respectively.

% \vspace{-0.2cm}
\subsubsection{Grasping the invisible}

The problem of “Grasping the Invisible” is proposed in \cite{yang2020deep}, where the target object is initially covered by other entities and is invisible to the cameras (see \cref{fig:invi_set}). Therefore, the robot needs to explore the workspace to find the target for subsequent grasping. Since we do not focus on the target not appearing in sight, we modify the original experimental settings in the following way for a fair comparison with the baseline methods. Before the action sequences start, a horizontal push towards the highest clutter is implemented to break the structure and investigate whether the target is buried inside.

Following the settings in \cite{yang2020deep}, the grasp is considered successful if the robot picks up the target within 15 motions. We execute the experiments by 10 runs on each case, where the results are shown in \cref{tab:invi_perf}. MASK-VPG achieves a success rate of 67.5\% with an average number of 11.6 motions, and GI performs better than MASK-VPG with a success rate of 85\% and an average number of 9.8 motions. Noticeable degradation in motion efficiency is observed for both MASK-VPG and GI in real-world settings compared with experiments in the simulated environment. Our GE-Grasp outperforms baseline methods by a large margin with a success rate of 97.5\% and only a 3.7 average number of motions due to the diverse action primitives and the spatial constraint in action candidate generation. Furthermore, we also rectified the definition of successful grasp that the robot picks up the target within 5 motions, which is the same as that in simulation experiments.
% which is applied in the experiments with the simulated environment.
In this more strict criterion, GE-Grasp achieves a 95\% success rate which is comparable with the performance in the simulated environment, indicating the strong generalization ability of GE-Grasp.

\begin{table}[t]
  \normalsize
  \centering
  \vspace{0.2cm}
  \begin{tabular}{lccc}
    \toprule
    Method & Success Rate (\%) & Motions\\
    \midrule
    MASK-VPG \cite{yang2020deep, zeng2018learning} & 67.5 & 11.6\\
    GI \cite{yang2020deep} & 85.0 & 9.8\\
    \textbf{GE-Grasp} & \textbf{97.5} & \textbf{3.7}\\
    \bottomrule
  \end{tabular}
  \caption{The success rate and the average number of motions for the task of ``Grasping the Invisible''.}
  \label{tab:invi_perf}
  \vspace{-0.2cm}
\end{table}

\subsubsection{Everyday objects}

\begin{figure}[t]
  \centering
  \includegraphics[width=1.0\linewidth]{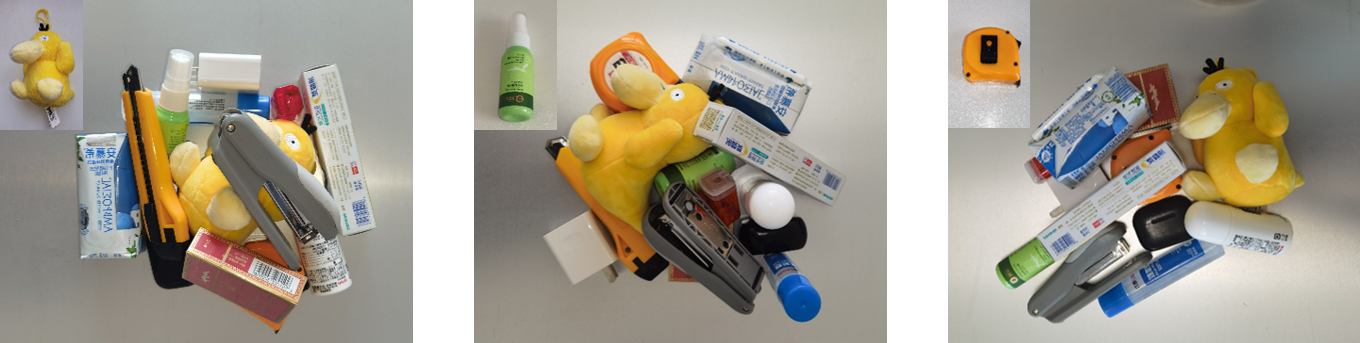}
  \caption{\textbf{Examples of the test cases containing everyday objects.} The clutter consists of 12 different everyday objects with one of which is assigned as the target each time.}
  \label{fig:everyday}
  \vspace{-0.6cm}
\end{figure}

To further investigate the generalization ability of our GE-Grasp, we change the target to novel objects in daily life while the model training still leverages the data collected in the simulated environment. The novel objects include stuffed toys, phone chargers, staplers, \etc as shown in \cref{fig:everyday}. The test scenarios contain 12 objects with different colors and shapes, and one of them is randomly selected as the target object with large parts occluded. The experiment is conducted by 10 runs for each case, and our GE-Grasp achieves a task success rate of 94.2\% with 4.1 motions on average. The results clearly show that our method can be generalized to novel objects in different sizes and shapes with the training blocks,  which verifies the practicality of GE-Grasp in realistic applications.

%-------------------------------------------------------------------------

\section{Conclusions and Discussion}

In this work, we have proposed the GE-Grasp framework for target grasping in dense clutter. The presented GE-Grasp leverages diverse action primitives for occluded object removal and employs a generator-evaluator architecture to avoid spatial collisions so that our approach efficiently grasps objects in dense clutter with promising success rates. Extensive experiments in both simulated and real-world environments have demonstrated the effectiveness and efficiency of our method. Moreover, GE-Grasp achieves comparable performance in the real world as that in the simulated environment, which indicates a strong generalization ability.

% \textbf{Limitations:} First, the primitive actions in our GE-Grasp are still strictly constrained (\ie fixed opening width of gripper and constant pushing distance), which degrades the ability to deal with different kinds of adversarial cases. Second, the application of top-down manipulation is seriously limited compared with 6-DOF manipulation, and modifying our GE-Grasp towards 6-DOF manipulation is a significant topic for the future work. Finally, our system performs actions without considering the dynamic changes in the workspace during execution, which may not be capable of handling unstable scenarios.

\section{Acknowledgment}

This work was supported in part by the National Natural Science Foundation of China under Grant U1813218 and Grant 62125603, and in part by a grant from the Beijing Academy of Artificial Intelligence (BAAI).

% {\small
% \bibliographystyle{ieee_fullname}
% \bibliography{ref}
% }

\end{document}